%% file: main.tex
\documentclass[sigconf]{acmart}

\AtBeginDocument{%
  }

\acmConference[IR-RAG@SIGIR'25]{Information Retrieval's Role in RAG Systems @ACM SIGIR Conference on Research and Development in Information Retrieval}{July 13--17,
  2025}{Padua, Italy}

\begin{document}

\title{TComQA: Extracting Temporal Commonsense from Text}


 \author{Lekshmi R Nair}
 \email{112001019@smail.iitpkd.ac.in}
 \affiliation{%
   \institution{Indian Institute of Technology}
   \city{Palakkad}
   \state{Kerala}
   \country{India}}

\author{Arun Sankar}
 \email{112001008@smail.iitpkd.ac.in  }
 \affiliation{%
   \institution{Indian Institute of Technology}
   \city{Palakkad}
  \state{Kerala}
   \country{India}}

 \author{Koninika Pal}
 \email{kpal.iitpkd.ac.in}
 \affiliation{%
   \institution{Indian Institute of Technology}
   \city{Palakkad}
   \state{Kerala}
   \country{India}
  }


\begin{abstract}
Understanding events necessitates grasping their temporal context, which is often not explicitly stated in natural language. For example, it is not a trivial task for a machine to infer that a museum tour may last for a few hours, but can not take months. Recent studies indicate that even advanced large language models (LLMs) struggle in generating text that require reasoning with temporal commonsense due to its infrequent explicit mention in text. Therefore, automatically mining temporal commonsense for events enables the creation of robust language models.
In this work, we investigate the capacity of LLMs to extract temporal commonsense from text and evaluate multiple experimental setups to assess their effectiveness. Here, we propose a temporal commonsense extraction pipeline that leverages LLMs to automatically mine temporal commonsense and use it to construct TComQA, a dataset derived from SAMSum and RealNews corpora. TComQA has been validated through crowdsourcing and achieves over 80\% precision in extracting temporal commonsense. The model trained with TComQA also outperforms an LLM fine-tuned on existing dataset of temporal question answering task.
\end{abstract}





\maketitle

\input{introduction.tex}

\input{method}

\input{experiments.tex}

\input{relatedworks.tex}

\input{conclusion.tex}

\bibliographystyle{ACM-Reference-Format}
\bibliography{main}


\end{document}

%% file: introduction.tex
\section{Introduction}

The automatic acquisition of knowledge has gained significant attention recently for developing AI systems capable of understanding and reasoning with human-like efficiency. 
Such a system needs information about typical properties of objects and concepts, which constitute commonsense knowledge that is rarely mentioned explicitly in natural language text. 
This work specifically focuses on automatically extracting temporal commonsense, which plays a crucial role in resolving time-related ambiguities and enabling contextually appropriate responses in conversation.

The major challenge in automatic acquisition of temporal commonsense is infrequent and imprecise mention of the explicit information related to the temporal arguments. 
For instance, when a user tells an AI assistant “Arrange me a babysitter for the vacation” vs
“Arrange me a babysitter for the evening”, the bot should understand that the user needs someone available for multiple days in the former case, while the latter only requires someone available for 4-5 hours. Additionally, events often span a range of plausible durations, making it difficult to associated with a definitive timeframe.
For example, the temporal context ``going on a trip" can be associated with all the  following valid temporal ranges, --a few days, or a few weeks, or a few months. Such variability introduces inherent ambiguity that can lead to a biased extraction models.

Many recent projects like Quasimode~\cite{DBLP:conf/cikm/RomeroRPPSW19}, WebChild~\cite{DBLP:conf/acl/TandonMW17}, ConceptNet~\cite{DBLP:series/tanlp/SpeerH13}, COMET~\cite{DBLP:conf/acl/BosselutRSMCC19}, etc. focus on the acquisition and creation of commonsense knowledge bases and aim to bridge the gap between existing knowledge about objects present in dominant knowledge bases like DBpedia, Freebase, Yago, etc., and their typical properties. 
Conversely, other works explore extracting temporal relations and temporal arguments from the text~\cite{DBLP:conf/semeval/StrotgenG10, DBLP:conf/naacl/VempalaBP18} but do not always capture all temporal properties for events. Some works focus on acquisition of physical properties~\cite{DBLP:conf/aaai/BagherinezhadHC16} like, size, weight, etc. and explore the distribution of quantitative attributes~\cite{DBLP:conf/acl/ElazarMRBR19} for specific events to harness commonsense knowledge. However, these efforts do not cover different types of temporal properties introduced by Zhou et al.~\cite{DBLP:conf/emnlp/ZhouKNR19} -- duration, frequency, stationary, ordering, and typical time. Moreover, recent literature in extracting temporal commonsense mainly reply on crowdsourcing, where annotators manually associate temporal ranges with events. This approach inherently limits the scalability of dataset creation and introduces biases based on individual annotators’ interpretations of temporal durations.  

In this work, we address this challenges by proposing an automated extraction pipeline that leverages LLMs to infer temporal commonsense for all five temporal properties, defined by Zhou et al.~\cite{DBLP:conf/emnlp/ZhouKNR19}, from short text. The pipeline first generates relevant questions for a target temporal properties and then utilize LLMs to extract corresponding answers from the context.
For example, given the context ``{\em Emma will be home soon and she will let Will know}", we would like to generate a question targeting temporal property {\em typical time}: ``{\em  When will Emma be home?}'' and find the answer ``6PM'' by exploiting the contextual information about typical workday routine. The contributions of this work are as follows:
\begin{itemize}
\item We propose a framework to extract temporal commonsense from the text by exploiting LLMs and introduce the TComQA dataset to enhance temporal Question-Answering task. 
\item We define two evaluation metrics to asses the quality and relevance of generated questions for a given context and specific temporal property.
    \item We conduct an extensive evaluation of each components of the proposed extraction pipeline. Furthermore, we validate the quality of the automatically generated TComQA dataset through crowdsourcing, demonstrating its high precision in extracting temporal commonsense. 
   \item The codes and dataset are available \href{https://anonymous.4open.science/r/TempQA-9D38/README.md}{\em here}.
\end{itemize}


%% file: method.tex
\section{Extraction of temporal commonsense from text}

\begin{figure*}[!t]
    \centering
    \includegraphics[trim={0 8cm 0 2cm},clip, width = 0.9\textwidth]{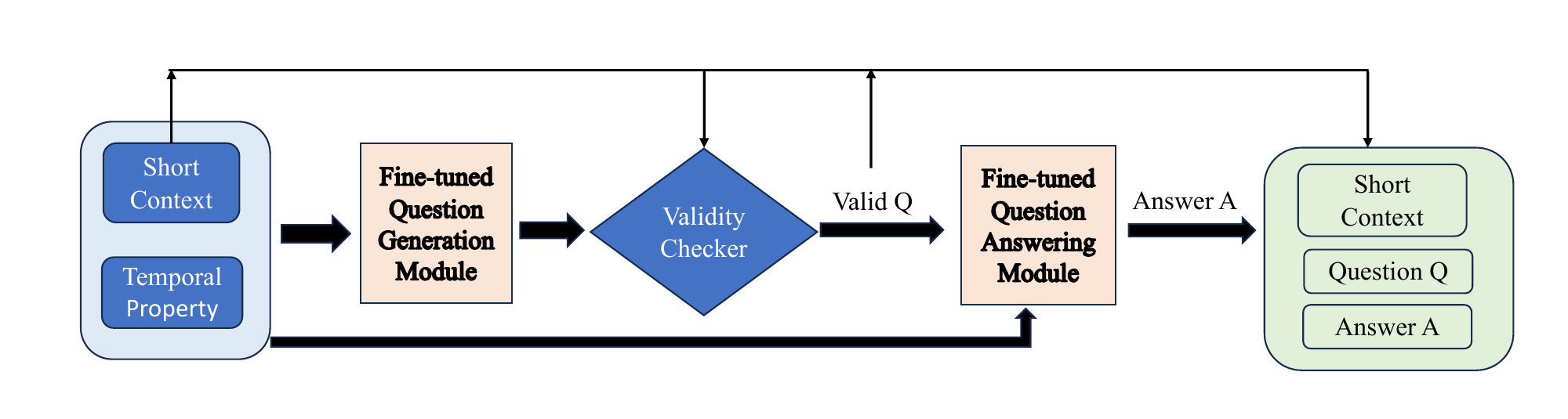}
    \caption{Overview of extraction of TComQA}
    \label{fig:flowchart}
\end{figure*}

The proposed framework for extracting temporal commonsense from text comprises three components: 1) a question generation module, 2) a question validity checker, and 3) an answer-generation module. 
Figure~\ref{fig:flowchart} illustrates the pipeline for proposed extraction framework that is used later to generate TComQA dataset. 

\subsection{Temporal Question Generation}
Extracting temporal commonsense from text requires accurately identifying the temporal events from the context and the associated temporal properties. 
Hence, the question generation module takes a short text with one of the five types of temporal commonsense reasoning, introduced by Zhou et al.~\cite{DBLP:conf/emnlp/ZhouKNR19}, to generate temporally relevant questions. We fine-tune the T5~\cite{DBLP:journals/corr/abs-1910-10683} architecture with the MCTACO~\cite{DBLP:conf/emnlp/ZhouKNR19} dataset to create this module. 
This dataset provides the type of temporal commonsense reasoning required to generate answers for a question. 

For instance, given a context:  ``{\em 
 Bartender asked for ID when Emily entered the bar and after checking, she was let in}" and temporal property: `{\em Typical Time}', we would like to generate a question: '{\em What is Emily's age?}'. 
 Answering such a question requires the temporal commonsense, and hence the answer becomes the extracted temporal commonsense for the event `entering the bar'.


\subsection{Question Validation}
The previous step generates questions for any given temporal properties. But, not all are appropriate for extracting the temporal value for the given context. 
Consider a context,`` Emma will be home soon and she will let Bob know'' and target temporal property: `frequency'. The question generator creates a relevant question for the event `coming home' matching the target temporal property: ``How often she is home after she is gone?'. 
But answering this question does not require any commonsense about the event, making it invalid for the commonsense extraction task. 
We observe that there is a contextual mismatch between the given context and the question when answering it for the target temporal property does not require any commonsense reasoning. 

Hence, we propose two metrics for validating the questions generated from the question generation module for a given target temporal property. Using these metrics, we first filter the valid questions generated by the question generation module and consider them for creating the TComQA dataset. 

\subsubsection{Validation Using Lexical Similarity.}
   The proposed lexical validator ($V_l$) measures the lexical similarity between the input context $C$ and the generated question $Q$. First, we filter noun, proper noun, and verb tokens from $C$ and $Q$ and lemmatize them. According to $V_l$, a question is considered valid if it contains a temporal marker ($T_m$) and there is at least one overlapping token between $C$ and $Q$. Temporal markers are tokens, such as "how often," "how long," "what time," "which year," "before," "afterward," etc. (an exhaustive list can be found in the code), obtained from the SPACY library. 

$$
   V_l(C,Q) =
    \begin{cases}
      1 &   \text{if}~|C \cap Q| >= 1 \wedge T_m \in Q  \\
      0 & \text{otherwise}
    \end{cases}    
$$

 \subsubsection{Validation using Semantic Similarity.} Since the lexical validator relies on exact token matching between a context and a question, it fails when the questions are generated from context using semantically similar words or phrases. 
Hence, we propose a semantic validator ($V_s$) that matches phrases between the question $Q$ and the context $C$. It first extracts noun and verb phrases from $C$ and $Q$, denoted as $C_{p}$ and $Q_{p}$ using the SPACY library. The semantic validator defines a question as valid if cosine similarity between embeddings of any phrase from $C$ and $Q$ satisfies the given similarity threshold and the temporal markers are present in the question. Clearly, as threshold $\theta$ tends to 1, this semantic validator becomes stricter.
\vspace{-1em}
$$
   V_s(C,Q) =
    \begin{cases}
      1 &   \text{if}~\exists_{i \in C_p, j \in Q_p} sim(i,j) >= \theta  \wedge T_m\in Q   \\
      0 & \text{otherwise}
    \end{cases}       
$$

\subsection{Generating Answers for Temporal Questions}
To extract the temporal commonsense, we finally need to find the answers for the valid questions generated by the question generation module. Here, we investigates different LLM-based answer generation models that takes a question with the associated context and the target temporal property and generates the typical value for the temporal property mentioned in the question. 

\textbf{Encoder-based Answer Generation Models.}
    The first approach consider BERT~\cite{DBLP:journals/corr/abs-1810-04805} as the underlying LLM and fine-tune it. As we need to generate numeric values for some of the temporal properties of an event, we also explore NumBERT~\cite{DBLP:journals/corr/abs-2010-05345}, which emphasizes numbers and represents them using a special token while learning the pre-trained model.
    As these are encoder-based approach, we append the MASK tokens of output size at the end of the input while finetuning these answer generation models. Clearly, this becomes a disadvantage of these models while finding the answer for a new question, because we need to know the length of the answer to mention the same number of [MASK] tokens in the input for predicting the answer. 
    
    \textbf{Decoder-based Answer Generation Model.} To overcome the drawback mentioned in the previous approaches, we adopt a decoder-based generative approach to find the answer. 
In this work, we fine-tuned Llama2-based~\cite{DBLP:journals/corr/abs-2307-09288} question-answering (QA) model for generating answers with dynamic length.

%% file: experiments.tex
\section{Evaluation}
\subsection{Datasets}
We use the MCTACO dataset~\cite{DBLP:conf/emnlp/ZhouKNR19} to fine-tune the underlying LLMs for creating both question-generation and answer-generation modules, as this dataset includes context, temporal questions, plausible answers, and the temporal properties needed to derive the answer from the context. It contains 3K short contexts with an average of 17.8 tokens per context. 

To create TComQA Dataset, we consider the SAMsum dataset~\cite{DBLP:journals/corr/abs-1911-12237}, a chat summarization dataset of 14K contexts with an average of 20.6 tokens per context, and the RealNews dataset~\cite{DBLP:conf/acl/ZhouNKR20} of 5K contexts with an average of 31.22 tokens per context.

\subsection{Experimental setup }

\subsubsection{Training setup for Question Generation Module}
We append a context with the temporal property that will be used for finding the answer for the associated question. It is then passed to the t5-large pre-trained model to train our question generation module, and the associated question is considered as output. 
We set max\_seq\_length, batch size, and epoch as 128, 8, and 10 respectively, to fine tune the t5-large pre-trained model to train our question-generation module. 

For training the encoder-based answer-generation models, we fine-tune the Bert\_base\_uncased and NumBERT using Adam optimizer with learning rate, batch size, and epochs set to 0.0001, 8, and 7 respectively. For fine-tuning NumBERT, we converted all numbers and numerical sequences according to the NumBERT directory. For the decoder-based model, we use Llama2 with 7 billion parameters. We fine-tune it using paged\_adamw\_32bit with a batch size of 4 and a learning rate of 2e-4.



\paragraph{Training setup for Question Answering Module}
For the encoder-based approach, we fine-tune Bert\_base\_uncased and NumBERT models using Adam optimizer with learning rate, batch size, and epochs set to 0.0001, 8, and 7 respectively. 
For fine-tuning NumBERT, we converted all numbers and numerical sequences according to the NumBERT directory. 

For the decoder-based model, we use Llama2 with 7 billion parameters. We trained it using paged\_adamw\_32bit with a batch size of 4 and a learning rate of 2e-4. 
Here we pass context, questions, temporal property to the model with associated answer to be generated. For example, the training data is formatted as
 $<s>$ [INST] context, question, temporal property [\/INST] answer $</s>$. Now, Llama2-based model generates comparatively long answers, whereas answers in our training data are comparatively short. For capturing that characteristics, we omit the generated sequence of tokens after $</s>$ while retrieving the answers from the model.

\subsection{Results and Discussion}

\paragraph{Evaluation of Question Generation Model.}
We consider 95 context-question pairs from MCTACO as test dataset and evaluate generated questions using semantic similarity. 
We achieved a moderate similarity score of 0.65 by considering the cosine similarity between the embedding of the generated questions and the groundtruth questions. Here, embedding of a question is calculated by averaging the token embedding from the SPACY library. The inherent variability of language leads to multiple valid questions for a given context, which is reflected in the relatively low similarity score. For example, given the context ``Our emphasis has been on having our potential clients know about us and deliver services to them, Dudovitz said. ", the model generates the question -- ``What did Dudovitz do after he spoke?" for the temporal property `Event Order'. Even though the generated question is valid and correctly align with the target temporal property, it is not semantically similar with the groupnth question for the context -- ``Do the clients learn about the company before becoming clients?". As a result, we reach a lower semantic similarity score for this task. Hence, we also manually evaluated the generated questions and found all the them were valid, matching the target temporal property for the input context. This happens as all these test dataset consider correct temporal property for the question generation task.

\paragraph{Evaluation of Question Validation Module}
For the MCTACO test samples, the proposed lexical validator $V_l$ confirms that 95\% of generated questions are valid. On the other hand, the semantic validator $V_s$ confirms the validity of 71\% questions with $\theta = 0.5$, 49\% with $\theta = 0.6$, and 37\% with $\theta = 0.7$. As expected, higher $\theta$ rejects more questions. While comparing the valid questions with the manually evaluated valid questions, we found that the lexical validator is more aligned with manual validation. For the SAMsum dataset, $V_l$ accepts 85\% of the generated questions and $V_s$ confirms the validity of 61\% questions with $\theta = 0.5$. 

In our extraction pipeline, we generate questions from a context for all five target temporal properties. So it is important to investigates further how well these two validators align with the human assessments. 
For this purpose, we sample 200 contexts from the SAMsum dataset and present the conext and the corresponding generated questions to assessors. 
Each context-question pair was evaluated by 5 judges, and they marked the question valid, invalid, or uncertain for the given context. We use majority votes to generate the groundtruth label for the generated question for a context. Based on the crowd-generated groundtruth, the $V_l$ achieves 89.7\% precision and 85.6\% recall and $V_s$ achieves 91.9\% precision with a lower recall of 69.5\%, as expected.

\paragraph{Evaluation of Question Answering Model.}
\begin{table}[t!]
    \centering
    \small
  \caption{Evaluation of answer generation module}
    \begin{tabular}{c|cc|cc|cc} 
        \hline
        &\multicolumn{2}{c}{BERT} &\multicolumn{2}{c}{NumBERT}& \multicolumn{2}{c}{Llama2}\\ \hline
        \textbf{ Prop.}  &DE & SS &DE & SS &DE &SS\\
        \hline
        Duration &   56.7\% &0.66 & 36.67\%& 0.42& 90\%& 0.37\\
        Typical Time &   60\% & 0.39&56.67\%& 0.38 & 86.67\%& 0.44\\
        Frequency &  56.7\% &0.63& 60\% &0.39& 86.67\% & 0.56\\
        Stationary & 40.0\% & 0.42 & 13\% & 0.40 & 83.34\% &0.38\\
        Event order &  16.67\% & 0.55 & 0\% & 0.38 & 76.67\% & 0.53\\ \hline
        \textbf{Total/Avg} & 46.02\%& 0.53 & 33.26\% & 0.39& 84.67\%& 0.47\\
        \hline
    \end{tabular}
    \label{QA_results}
\end{table}
Table~\ref{QA_results} presents the evaluation of all three answer-generation models tested on 150 question-answer pairs, 30 from each temporal types, from the MCTACO test dataset. in this experimental setup, generated answer can have different surface forms than the groundtruth answer for a question. 
For instance, 'each year' is similar to `every year' or `rarely' is semantically similar to `once' for temporal property {\em frequency}. Furthermore, there are multiple valid answers for the temporal property `typical time', e.g., the question, ``what time of the day was the storm", can have multiple valid answers, such as 3 PM and 4 PM. Hence, we consider evaluating the answer generation module using the semantic similarity (SS) between the embeddings of generated answers and groundtruth answers. We can see both BERT and Llama2-based models reach moderate semantic similarity for answer generation task. But due to answer variability mentioned before, semantic similarity (SS) does not directly reflect the preciseness of the model on correct answer generation. Hence, we validate the answer by domain experts (DE) given the reference of groundtruth answers, and we found Llama2-based generative model outperforms others, reaching 84\% precision in answer generation. 

\begin{table}[t]
    \centering
    \small
    \caption{Evaluation of TComQA answers using crowdsourcing}
    \begin{tabular}{c|c|c|c|c}
        \hline
        {Duration}  & {Typical Time}  & {Frequency} &{stationary} & Event order\\
        \hline
        76.67\% &  90\% & 72.41\%& 77.77\% & 89.47\%\\
        \hline
    \end{tabular}
    \label{Crowd_results}
\end{table}

\subsection{Generation and Evaluation of TComQA} 
Based on the evaluation, we consider the best-performing models for three components of our extraction pipeline. Therefore, we use the T5-based fine-tuned question generation module, then validates the generated questions using the lexical validator, and finally produces the answer using the Llama-2-based fine-tuned answer-generation model.  Using the SAMsum and RealNews, we generated a total of 17K valid Question-answer pairs with associated temporal properties. We sampled 150 of them, 30 for each temporal property, and use crowdsourcing to evaluate them as no groundtruth answers are available.
Each answer was assessed by 5 judges and labeled as valid, invalid, or uncertain. Groundtruth labels are generated using majority vote from corwedsourced data. Table~\ref{Crowd_results} records the precision of the extracted answer for each temporal types, achieving overall 81.26\% precision. 

Further, we fine-tuned the answer-generation model with the TComQA using Llama2 and evaluated it on the MCTACO test set. We consider the semantic similarity between the groundtruth data and generated answer as the evaluation metric. Table~\ref{semantic similarity} summarizes the performance of the fine-tuned model for all five temporal types. The evaluation shows improvement in the the answer generation task compared to the model created using the MCTACO dataset(cf. Table~\ref{QA_results} last column). Specifically for duration, typical time, and frequency, it reaches 0.52, 0.56, and 0.59 semantic score respectively. This reflects that TComQA helps in temporal commonsense inference task.

\begin{table}[t]
    \centering
    \small
    \caption{Evaluation of QA model fine-tuned with TComQA}
    \begin{tabular}{c|c|c|c|c|c}
        \hline
        Metrics & {Duration}  & {Typical Time}  & {Frequency} &{stationary} & order\\
        \hline
       SS&  0.52 &  0.56 & 0.59& 0.41 & 0.41\\
       DE &86.7\% &  86.7\% & 93\%& 86.7\% & 76.6\%\\
        \hline
    \end{tabular}
    \label{semantic similarity}
\end{table}

%% file: relatedworks.tex
\section{Related Works}
Here, we specifically focus on the recent works related to extraction, reasoning, and inference of temporal commonsense. Many studies aim to precisely structure temporal events, specifically finding the event duration~\cite{DBLP:conf/acl/RothWN18, DBLP:conf/naacl/VempalaBP18}, extraction of relative timeline ~\cite{DBLP:conf/emnlp/LeeuwenbergM18}, and normalization of temporal expressions~\cite{DBLP:conf/acl/LeeADZ14} for temporal events. 
Some works focus on event ordering, considering causal and temporal relations between events~\cite{DBLP:conf/acl/RothWNF18, DBLP:conf/coling/MirzaT16}. Zhou et al.~\cite{DBLP:conf/emnlp/ZhouKNR19} categorize the temporal properties of events into five types of temporal commonsense and create the MCTACO dataset by enhancing crowdsourced question-answer pairs to cover all five types. 
Based on WikiHow pages, Lynden et al~\cite{DBLP:conf/cikm/LyndenYKJMYLD23} extracted a large dataset on the duration of temporal events using crowdsourcing, demonstrating the effectiveness of inference tasks using LLMs.

With the advancement and availability of pre-trained LLMs, many studies have explored their power for commonsense inference tasks~\cite{DBLP:conf/ijcnlp/KimuraPK22,DBLP:conf/emnlp/JainSA0JD23, DBLP:conf/acl/ZhouNKR20}. Jain et al.~\cite{ DBLP:conf/emnlp/JainSA0JD23} present a survey of different LLMs on temporal inference task, discussing their limitations and potential future directions. Qin et al.~\cite{DBLP:conf/acl/QinGUHCF20} investigate LLMs on temporal reasoning over discourse and curated a crowdsoursced dialogue dataset for this purpose. They found that LLMs are less efficient at correlating temporal contexts in conversation compared to humans. Zhou et al.~\cite{DBLP:conf/acl/ZhouNKR20} propose a pattern-based approach to extract temporal arguments covering three different temporal properties and use sequential language modelling to generate time-specific language model that enhances the performance of temporal commonsense inference task over BERT. 
Murata et al.~\cite{DBLP:conf/coling/MurataK24} extend COMET by appending time between events that further improve the commonsense inference task and provide better embedding representations for events. In this work, to overcome the scaling issue of crowdsourcing in temporal commonsense acquisition task, we propose an extraction pipeline for temporal commonsense by leveraging the power of LLMs with the prompts about temporal property.

%% file: conclusion.tex
\section{Conclusion}
We developed a temporal commonsense extraction pipeline using different large language models (LLMs). The framework identifies the temporal event and provides appropriate questions for a temporal property along with along with the plausible answers for the temporal event. With this framework, we created temporal commonsense QA dataset, TcomQA, from two distinct real-world datasets. Our approach eliminates the dependency on manual annotation, enabling large-scale, consistent, and efficient generation of temporal Automatically generated TComQA dataset overcomes the scaling issue in commonsense dataset generation. The evaluation of TcomQA demonstrated that our extraction pipeline successfully captures temporal commonsense from text with a precision score exceeding 80\% and improve the temporal Question-answering task.